\documentclass{wscpaperproc}
\usepackage{latexsym}
\usepackage{graphicx}
\usepackage{mathptmx}
\usepackage[T1]{fontenc}

\usepackage{amsmath}
\usepackage{amsfonts}
\usepackage{amssymb}
\usepackage{amsbsy}
\usepackage{amsthm}
\usepackage{subfig}
\usepackage{graphicx}
\usepackage{soul}
\usepackage{colortbl} 
\usepackage{array}    
\usepackage{booktabs}
\usepackage{subcaption} 
\usepackage[normalem]{ulem}
\usepackage{multirow}
\usepackage{wrapfig} 
\usepackage[table]{xcolor}
\usepackage{colortbl}

\newcommand{\fullname}{\textbf{\textit{\underline{Sim}}}ulation-to-\textbf{\textit{\underline{Dec}}}ision}
\newcommand{\shortname}{Sim-to-Dec}
\usepackage[pdftex,colorlinks=true,urlcolor=blue,citecolor=black,anchorcolor=black,linkcolor=black]{hyperref}

\newtheoremstyle{wsc}
{3pt}
{3pt}
{}
{}
{\bf}
{}
{.5em}
{}

\theoremstyle{wsc}

\begin{document}

%
%

\pagestyle{fancyplain}

\thispagestyle{plain}
\firstPageHead{}

\chead{\fancyplain{}{\itshape Bai, Wang, Gong, Wang, Ying, Chen, and Fu}}

\rhead{}
\cfoot{}
\renewcommand{\headrulewidth}{0pt} 

\makeatletter
\let\@internalcite\cite
\def\cite{\def\@citeseppen{-1000}%
    \def\@cite##1##2{(##1\if@tempswa , ##2\fi)}%
    \def\citeauthoryear##1##2##3{##1 ##3}\@internalcite}
\def\citeNP{\def\@citeseppen{-1000}%
    \def\@cite##1##2{##1\if@tempswa , ##2\fi}%
    \def\citeauthoryear##1##2##3{##1 ##3}\@internalcite}
\def\citeN{\def\@citeseppen{-1000}%
    \def\@cite##1##2{##1\if@tempswa, ##2)\else{}\fi}%
    \def\citeauthoryear##1##2##3{##1 (##3)}\@citedata}
\def\citeA{\def\@citeseppen{-1000}%
    \def\@cite##1##2{(##1\if@tempswa , ##2\fi)}%
    \def\citeauthoryear##1##2##3{##1}\@internalcite}
\def\citeANP{\def\@citeseppen{-1000}%
    \def\@cite##1##2{##1\if@tempswa , ##2\fi}%
    \def\citeauthoryear##1##2##3{##1}\@internalcite}
\def\shortcite{\def\@citeseppen{-1000}%
    \def\@cite##1##2{(##1\if@tempswa , ##2\fi)}%
    \def\citeauthoryear##1##2##3{##2 ##3}\@internalcite}
\def\shortciteNP{\def\@citeseppen{-1000}%
    \def\@cite##1##2{##1\if@tempswa , ##2\fi}%
    \def\citeauthoryear##1##2##3{##2 ##3}\@internalcite}
\def\shortciteN{\def\@citeseppen{-1000}%
    \def\@cite##1##2{##1\if@tempswa, ##2\else{}\fi}%
    \def\citeauthoryear##1##2##3{##2 (##3)}\@citedata}
\def\shortciteA{\def\@citeseppen{-1000}%
    \def\@cite##1##2{(##1\if@tempswa , ##2\fi)}%
    \def\citeauthoryear##1##2##3{##2}\@internalcite}
\def\shortciteANP{\def\@citeseppen{-1000}%
    \def\@cite##1##2{##1\if@tempswa , ##2\fi}%
    \def\citeauthoryear##1##2##3{##2}\@internalcite}
\def\citeyear{\def\@citeseppen{-1000}%
    \def\@cite##1##2{(##1\if@tempswa , ##2\fi)}%
    \def\citeauthoryear##1##2##3{##3}\@citedata}
\def\citeyearNP{\def\@citeseppen{-1000}%
    \def\@cite##1##2{##1\if@tempswa , ##2\fi}%
    \def\citeauthoryear##1##2##3{##3}\@citedata}
%
%
%
\def\@citedata{%
    \@ifnextchar [{\@tempswatrue\@citedatax}%
                  {\@tempswafalse\@citedatax[]}%
}

\def\@citedatax[#1]#2{%
\if@filesw\immediate\write\@auxout{\string\citation{#2}}\fi%
  \def\@citea{}\@cite{\@for\@citeb:=#2\do%
    {\@citea\def\@citea{, }\@ifundefined
       {b@\@citeb}{{\bf ?}%
       \@warning{Citation `\@citeb' on page \thepage \space undefined}}%
{\csname b@\@citeb\endcsname}}}{#1}}%

%
\def\@citex[#1]#2{%
\if@filesw\immediate\write\@auxout{\string\citation{#2}}\fi%
  \def\@citea{}\@cite{\@for\@citeb:=#2\do%
    {\@citea\def\@citea{; }\@ifundefined
       {b@\@citeb}{{\bf ?}%
       \@warning{Citation `\@citeb' on page \thepage \space undefined}}%
{\csname b@\@citeb\endcsname}}}{#1}}%

%
\def\@biblabel#1{}
\makeatother



\newdimen\bibindent
\bibindent=0.0em
\def\thebibliography#1{\section*{\refname}\list
   {}{\settowidth\labelwidth{[#1]}
   \leftmargin\parindent
   \itemindent -\parindent
   \listparindent \itemindent
   \itemsep 0pt
   \parsep 0pt}
   \def\newblock{}
   \sloppy
   \sfcode`\.=1000\relax}


\setlength{\baselineskip}{12.7pt}

\title{Supply Chain Optimization via Generative Simulation and Iterative Decision Policies}

\author{\begin{center}Haoyue Bai\textsuperscript{1}, Haoyu Wang\textsuperscript{2}, Nanxu Gong\textsuperscript{1}, Xinyuan Wang\textsuperscript{1}, Wangyang Ying\textsuperscript{1}, Haifeng Chen\textsuperscript{2}, and Yanjie Fu\textsuperscript{1}\\
[11pt]
\textsuperscript{1}School of Computing and Augmented Intelligence, Arizona State University, AZ, USA\\
\textsuperscript{2}Data Science and System Security, NEC Labs America, NJ, USA
\end{center}
}

\maketitle

\section*{ABSTRACT}
High responsiveness and economic efficiency are critical objectives in supply chain transportation, both of which are influenced by strategic decisions on shipping mode. An integrated framework combining an efficient simulator with an intelligent decision-making algorithm can provide an observable, low-risk environment for transportation strategy design. An ideal simulation-decision framework must (1) generalize effectively across various settings, (2) reflect fine-grained transportation dynamics, (3) integrate historical experience with predictive insights, and (4) maintain tight integration between simulation feedback and policy refinement. We propose \shortname\ framework to satisfy these requirements. Specifically, \shortname\ consists of a generative simulation module, which leverages autoregressive modeling to simulate continuous state changes, reducing dependence on handcrafted domain-specific rules and enhancing robustness against data fluctuations; and a history–future dual-aware decision model, refined iteratively through end-to-end optimization with simulator interactions. Extensive experiments conducted on three real-world datasets demonstrate that \shortname\ significantly improves timely delivery rates and profit.

\section{Introduction}

Efficient transportation plays a central role in supply chains across commerce, finance, and agriculture industries, directly supporting logistics execution, demand fulfillment, and resource coordination~\cite{e-commerce,financial,agriculture}.  
A well-functioning supply chain transportation system must strike a balance between high responsiveness and economic efficiency, where achieving this balance largely depends on strategic transportation decisions~\cite{TripleR,res}.  
Transportation decisions, such as selecting for each order between air, rail, or maritime shipping, involve inherent trade-offs between delivery speed and cost.  
Faster modes improve responsiveness but incur higher expenses, while slower options reduce costs at the risk of delay.  
Optimizing such decisions is essential for improving responsiveness and efficiency under dynamic supply chain conditions.


Simulation has long served as a powerful tool for evaluating supply chain transportation strategies before real-world deployment~\cite{sim4pt1,sim4pt2}. A promising direction is to couple simulation with intelligent decision-making, enabling continuous optimization in a low-risk and adaptive environment.
Several research directions are intrinsically relevant to the development of integrated simulation–decision frameworks. Traditional simulation techniques, such as discrete event simulation~\cite{discreteEvent} and Monte Carlo methods~\cite{monteCarlo}, provide detailed modeling of logistics processes but often require extensive expert knowledge and manual parameter tuning, which can limit their scalability across diverse operational settings. Reinforcement learning (RL) approaches~\cite{rl_intro} enable adaptive policy optimization through interaction with the environment. However, when applied to complex supply chain systems, RL methods may face practical challenges such as high computational costs and sensitivity to objective changes~\cite{RL_cost}. They often treat the environment as a black box or assume fixed, coarse-grained dynamics, which limits their ability to model fine-grained state transitions or adapt to real-world disruptions such as demand fluctuations, policy shifts, and capacity constraints~\cite{RL_envstate}. In addition, many existing methods loosely couple simulation and decision-making: traditional simulators are typically used as post-hoc evaluators~\cite{simanddm,simanddm2} without actively guiding the decision process.

To effectively support transportation strategy design, we argue that an ideal simulation–decision framework should satisfy four key criteria.  
(1) \textit{Generalizability} — the framework should be data-driven and easily adaptable to various supply chain settings without relying on handcrafted, domain-specific modeling.  
(2) \textit{Dynamic fidelity} — it should capture fine-grained transportation dynamics, including continuous and interdependent state transitions that evolve over time.  
(3) \textit{Experience-forecast integration} — the decision model should leverage both historical operational data and future-oriented predictions to formulate informed strategies.  
(4) \textit{Tight simulation–decision coupling} — the simulator and the decision model should interact iteratively, enabling the simulation to guide the decision-making rather than operating in isolation.

\begin{wrapfigure}{r}{0.38\textwidth}  
    \centering
    \includegraphics[width=0.38\textwidth]{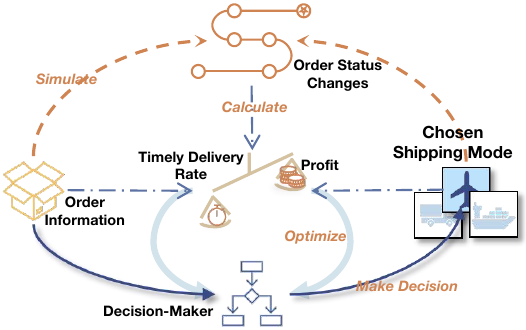}
    \caption{\small{Proposed Framework}}
    \label{fig:intro}
\end{wrapfigure}
To meet these criteria and support adaptive, efficient transportation strategy design, we propose \fullname\ (\shortname), a novel framework that unifies generative simulation with end-to-end decision optimization.
As illustrated in Figure~\ref{fig:intro}, the framework operates in three stages. First, the decision model selects an appropriate shipping mode for each order based on its attributes. Then, the simulator predicts the order’s status evolution conditioned on the order details and the chosen mode. Finally, the system evaluates the strategy using key metrics such as profit and on-time delivery rate based on the simulated states.
\shortname~consists of two tightly integrated components:  
(1) \textit{Generative simulation for scalable and data-driven modeling.} The simulator learns transportation dynamics directly from historical data, eliminating the need for manual rule engineering. It is implemented as an autoregressive model that sequentially predicts order states across transportation stages, capturing both short-term fluctuations and long-term trends. This enables fine-grained simulation of continuous and interdependent logistics behaviors, supporting generalization across diverse supply chain settings.  
(2) \textit{End-to-end decision-making with independent yet coupled learning.} The decision model is parameterized independently from the simulator, allowing it to flexibly adapt to evolving objectives and constraints. It interacts iteratively with the simulator in a low-risk virtual environment, continuously refining transportation strategies through simulated feedback. To support experience-forecast integration, the model combines two complementary perspectives: historical execution data is used to estimate expected outcomes of shipping modes, while a value network predicts future rewards. These sources jointly guide policy optimization, enabling decisions that are both historically grounded and forward-looking.
By integrating these components, \shortname~forms a closed feedback loop: the generative simulator infers dynamic logistics conditions, while the decision model actively learns and adapts strategies. This architecture directly addresses the four key design goals outlined earlier—generalizability, dynamic fidelity, experience-forecast integration, and tight simulation–decision coupling—making \shortname~well-suited for the complexity and uncertainty of real-world supply chain transportation.

\textbf{Key Contributions:} This work makes the following contributions to supply chain transportation optimization: (1) \textbf{Framework:} We propose a novel simulation-decision framework that integrates generative simulation with iterative decision-making, providing a flexible and adaptive approach for optimizing transportation strategies in the supply chain. (2) \textbf{Decision Optimization:} Our method combines insights from historical data with future reward estimation to guide decision-making, achieving a balance between leveraging experience and adapting to evolving conditions. (3) \textbf{Empirical Validation:} We validate the proposed framework through extensive experiments on three real-world supply chain datasets and a live transportation system, demonstrating significant improvements in transportation performance, responsiveness, and decision robustness. The code is available at \url{https://github.com/HaoyueBai98/Sim-to-Decision}.

This paper is organized as follows. Section 2 provides background on shipping mode management in supply chains and formally defines the problem. Section 3 introduces our Sim-to-Dec framework, detailing its two core components: a generative simulator and a decision-making model. Section 4 presents our experimental setup and results, including comparisons with baselines and ablation studies. Section 5 reviews related work on simulation and decision-making in supply chain optimization. We conclude with a discussion of future directions.

\section{Related Work}

Simulation-based methods are widely used to evaluate supply chain transportation strategies under uncertainty. Traditional techniques include discrete event simulation (DES)~\cite{discreteEvent}, system dynamics (SD)~\cite{sd}, and agent-based modeling (ABM), each capturing different aspects of system characteristics and behaviors. DES is particularly suitable for modeling operational-level disruptions, SD provides insights into system-wide feedback-driven behaviors, and ABM effectively models decentralized agent interactions. Hybrid approaches that combine these paradigms have also been explored to leverage their complementary modeling strengths~\cite{anylogicHybrid}.
Probabilistic methods such as Monte Carlo simulation~\cite{monteCarlo} and Markov-based modeling~\cite{hosseini2019review} are also commonly employed, particularly for quantifying variability and modeling stochastic transitions or cascading failures~\cite{dixit2020assessment}. These traditional simulations offer strong interpretability and transparency, but typically require extensive expert knowledge and rule-based parameterization, which significantly limits their adaptability to dynamic and evolving operational conditions.

On the decision-making side, classical optimization methods (e.g., linear programming~\cite{LP,LP2}) have long been used to derive cost-effective policies. More recently, reinforcement learning (RL) has emerged as a promising alternative~\cite{Rl-based_survey}, allowing agents to learn adaptive strategies through continuous environmental interaction. However, RL approaches are often sample inefficient, computationally expensive, and assume black-box environments with coarse dynamics~\cite{adobor2020supply,RL_cost,RL_envstate}, which significantly undermines their applicability in complex, fine-grained supply chains.
To address these limitations, recent research has explored integrating simulation and decision-making into unified frameworks. For example, Correa-Martinez and Seck~\cite{correa2023generic,bai2025brownian} and others have proposed simulation-driven policy optimization or digital twin architectures~\cite{digitaltwin}, enabling continuous improvement of decision strategies under uncertainty. However, many of these frameworks still treat simulation as a passive evaluator rather than an active participant in iterative decision refinement. The coupling between simulation and policy learning remains loose~\cite{simanddm,simanddm2}, and simulation fidelity is often limited to simplified dynamics or offline scenario replay, which reduces generalizability.
Our work contributes to this growing literature by proposing a tightly integrated simulation–decision framework that directly overcomes these limitations.

\section{Background and Problem Statement}



\noindent
\textbf{Challenges of Shipping Mode Management \& Order Representation.}  
In supply chain transportation, selecting an appropriate shipping mode is a fundamental decision that directly affects both responsiveness and cost efficiency. 
Expedited modes such as air freight improve delivery speed but incur higher costs, while slower options like maritime or rail transport are more economical but risk significant delays. 
Thus, one of the key challenges in shipping mode management is to strategically balance timely delivery rates and profits through mode selection.
To support this decision process, each order in the supply chain is represented by three distinct groups of attributes:
\ul{(1) Order information} related attributes, denoted by $F^I$, are a set of attributes observed and collected at the time of order placement. 
The order information is inherent attributes that are not affected by decisions and will not change over time, denoted by $F^I = \{f^I_1, f^I_2, \dots, f^I_{|F^I|}\}$, where $|F^I|$ is the number of order information related attributes, for instance, the origin, destination, product type, quantity, and required delivery time of an order.
\ul{(2) Shipping modes}, denoted by ($F^D$), are a set of candidate shipping modes (in our experiments, we has four shipping modes) for an order, denoted by $F^D = \{f^D_1, f^D_2, \dots, f^D_{|F^D|}\}$ .
Each $f^D_d \in F^D $ is a specific shipping mode chosen from a predefined set of options, such as air, maritime, or ground transport.
\ul{(3) Order states} related attributes, denoted by $F^E$, are a set of attributes describing the changes or evolutions of an order after a shipping mode is selected since an order has dynamics and its state changes. 
We define three order state attributes: 
i) a binary variable indicating whether the order has a risk of delay, denoted by $f^E_{\text{risk}}$, for instance, high risk or low risk. 
ii) a categorical variable indicating the number of days required for delivery, $f^E_{\text{time}}$, for instance,  1, 2, 3, or 4 days. 
iii) a binary variable indicating whether the order is ultimately delivered on time, denoted by $f^E_{\text{status}}$.  

\noindent
\textbf{Simulation–Decision Integration for Shipping Strategy Optimization.} 
The goal of shipping mode management is to jointly optimize two key metrics: the timely delivery rate ($T^{\text{Timely}}$) and profit ($T^{\text{Profit}}$), which are often in conflict. Direct deployment of optimization strategies in real-world systems is infeasible due to operational risks and constraints. To address this, we propose \shortname, a closed-loop simulation–decision framework that enables risk-free evaluation and refinement of transportation strategies using historical data.

\shortname~consists of two tightly coupled components: a generative simulator and a policy-based decision maker. The simulator, denoted as $\mathcal{S}$, learns fine-grained transportation dynamics from historical data and simulates the impact of shipping decisions. The decision maker, $\mathcal{M}$, learns a policy to select optimal shipping modes for individual orders, based on their information features $F^I_n$. Specifically, for the $n$-th order:

\begin{equation}
\hat{f}^D_{d,n}= \mathcal{M}(F^I_n; \Phi),
\end{equation}

where $\Phi$ is the parameter set of the decision model. The selected shipping mode $\hat{f}^D_{d,n}$ is then passed into the simulator to generate the predicted order state $\hat{F}_n^E$:

\begin{equation}
\hat{F}_n^E = \mathcal{S}(F^I_n, \hat{f}_{d,n}^D; \Theta),
\end{equation}

where $\Theta$ denotes the simulator's parameters.
The simulation provides feedback on expected outcomes, which is used to iteratively refine the decision policy. The objective is to maximize both $T^{\text{Timely}}$ and $T^{\text{Profit}}$ over a batch of $N$ orders. 
After preprocessing, all necessary attributes—including order details, candidate modes, and outcomes—are available from real datasets to support both training and evaluation.

\section{The \shortname~Approach}
\shortname\ consists of two integrated components: a \textit{generative simulator} and a \textit{decision-maker}. 
The generative simulator models the dynamics of a supply chain system by learning from historical order data and generating how the supply chain system changes when the shipping modes of orders are chosen. 
The decision-maker interacts with the generative simulator to iteratively optimize shipping mode selection, ensuring a balance between timely delivery and cost efficiency. Through continuous simulation feedback, the decision-maker dynamically refines decision policies to adapt to changing supply chain conditions.

\subsection{Generative Simulator via Deep Generative AI}
\begin{figure*}[ht]
\centering
\includegraphics[width = 1\textwidth]{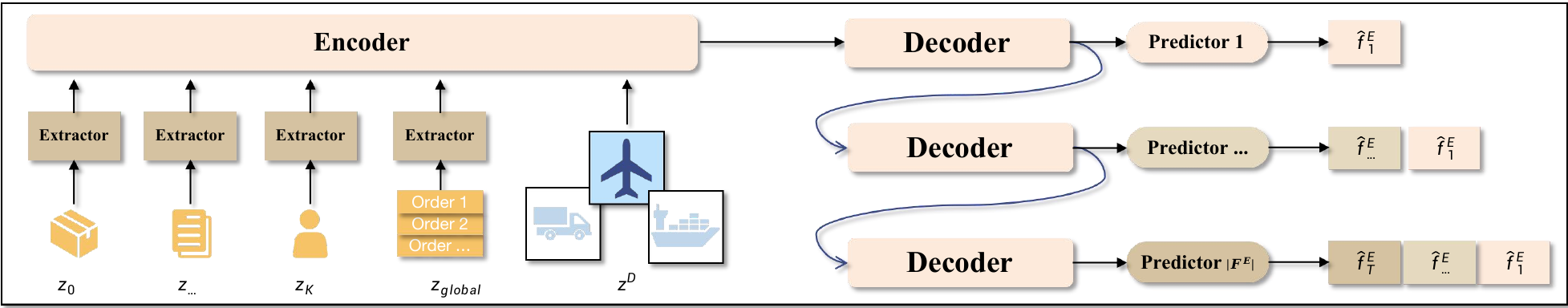}
\caption{\small{Simulator}}
\label{fig:sim}
\end{figure*}

Traditional simulation approaches in supply chain shipping mode management heavily rely on expert-driven models with predefined heuristics, making them rigid and difficult to adapt to evolving market conditions. 
To address these limitations, we propose a \textit{generative simulator} that models simulation as a generative AI task by learning directly from offline historical data. This approach eliminates the need for extensive expert knowledge while allowing for adaptive, high-fidelity simulations.

\noindent
\textbf{Processing Order Information.}  
To simulate order dynamics, we categorize \textit{order information} ($F^I$) into multiple groups, each representing distinct aspects of an order’s attributes:
\ul{(1) Product-Related Attributes $F^{I,p}$:} characteristics of a shipped item, such as weight and product category.
\ul{(2) Customer-Related Attributes $F^{I,c}$:} characteristics of an order recipient, including location, priority status, and historical purchase behavior.
\ul{(3) Shipping-Related Attributes $F^{I,s}$:} order-specific constraints, such as scheduled delivery time, shipping restrictions, and previous shipping history.
\ul{(4) Order-Related Attributes $F^{I,o}$:} statistical order attributes, including order type, order placement time, and payment method.

Since individual shipping events are interconnected, we introduce a global order group $F^{I,\text{global}}$ as an additional attribute group that captures dependencies among multiple orders processed within the same time frame. This global representation is constructed by \textit{pooling} information from all orders in a given batch, ensuring that system-wide effects:
\begin{equation}
F^{I,\text{global}} = \text{Pool}(\{F^I_n\}_{n=1}^{N}),
\end{equation}
where $\text{Pool}(\cdot)$ aggregates the attribute representations of all $N$ orders in a batch, capturing an overall system-wide summary. 
Each attribute group, including the global representation, is processed using a separate linear transformation layer:
\begin{equation}
\mathbf{z}^k_n = \mathbf{W}^k F^{I,k}_n + \mathbf{b}^k, \quad \forall k \in \{p,c,s,o, \text{global} \},
\end{equation}
where $k$ denotes different attribute groups, $\mathbf{W}^k$ and $\mathbf{b}^k$ the learnable vector and $\mathbf{z}^k_n$ is the representation of group $k$.
This transformation ensures that all attribute groups, including the global representation, are mapped into a consistent latent space before being fed into the downstream model. The processed attribute embeddings are then treated as input tokens for the encoder, allowing it to simultaneously capture both individual shipment details and system-wide patterns.

\noindent\textbf{Processing Shipping Mode.}
To simulate system dynamics conditioned on shipping decisions, we embed the selected shipping mode $f_n^{D}$ into a latent representation $\mathbf{z}_n^D$. 
We implement this via an embedding lookup:
\begin{equation}
\mathbf{z}_n^D = \text{Embedding}(f^{D}_n), \quad f^{D}_n \in F^D,
\end{equation}
where the embedding table is initialized using samples from a normal distribution.


\noindent\textbf{Fusing and Encoding.}
We utilize the Long Short-Term Memory~(LSTM) model as an encoder to encode the transformed representations of both order information and shipping mode into a fused embedding. Formally, the encoding process is given by:
\begin{equation}
\mathbf{z}_n = \text{LSTM}(\mathbf{z}^p_n, \mathbf{z}^c_n, \mathbf{z}^s_n, \mathbf{z}^o_n, \mathbf{z}^{global}_n, \mathbf{z}_n^D).
\end{equation}

\noindent
\textbf{Generating Order Status Changes.}  
Order status changes describe how orders in the system change when the decision maker chooses different shipping modes. Understanding what will happen after taking a decision can enable us to measure the utility of a decision and adjust the policies of the decision maker. 
We see the simulation as a task of generating evolutionary attributes given the necessary information, i.e., order information and shipping mode. 
In particular,  we utilize the LSTM decoder to autoregressively generate evolutionary attributes. Using the fused embedding $\mathbf{z}_n$, the decoder sequentially generates the latent representations of changes in order status. Formally, we generate the embedding $\mathbf{z}^E_{e,n}$  of the $e^{th}$ order status attribute of the order $n$ by: 
\begin{equation}
\mathbf{z}^E_{e,n} = \text{LSTM}(\mathbf{z}_n, \mathbf{z}^E_{<e,n}),
\end{equation}
where $\mathbf{z}^E_{<e,n}$ denotes the embedding of the order status attribute from $1^{st}$ to $({e-1})^{th}$.
Finally, a dedicated predictor is used to predict the value of a specific order status attribute. Formally, given the embedding $\mathbf{z}^E_{e,n}$ of the $e^{th}$ evolutionary attribute, let $\mathcal{P}_e$ be the dedicated predictor of $e^{th}$ order status attribute, the generated/simulated value $\hat{f}^E_{e,n}$  of the $e^{th}$ order status attribute is given by:
\begin{equation}
\hat{f}^E_{e,n} = \mathcal{P}_e(\mathbf{z}^E_{e,n}).
\end{equation}

\noindent
\textbf{Final Optimization Objective of Simulator.}
The simulator's ability to accurately reflect real-world dynamics is essential for reliable analysis and informed decision-making. To ensure fidelity, we leverage real-world historical data as ground truth to train the model. Each order status is optimized by minimizing the following objective function:
\begin{equation}
\mathcal{L}_{\mathcal{S}} = \sum_{e=1}^{T}\sum_{n=1}^{N} \left( \hat{f}^E_{e,n} - f^E_{e,n} \right)^2,
\end{equation}
where $f^E_{e,n}$ is the true value of the $e^{th}$ order status attribute of order $n$, and $\hat{f}^E_{e,n}$ is the predicted value.
By minimizing this loss function, the simulator learns to accurately reflect the dynamics of the system, providing valuable insights and guidance for decision-makers.

\subsection{Decision Maker via Policy Neural Networks}
\begin{figure*}[ht]
\centering
\includegraphics[width = 1\textwidth]{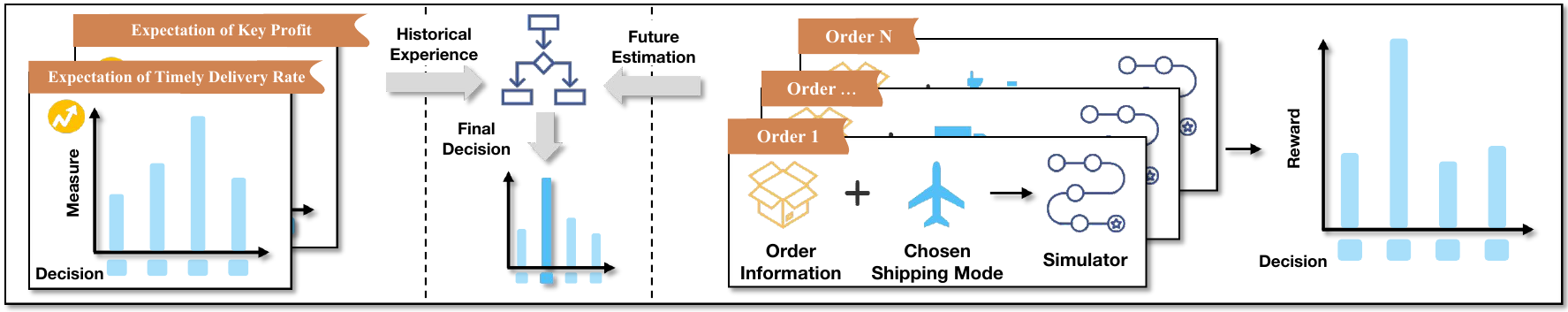}
\caption{\small{Decision-Maker}}
\label{fig:dm}
\end{figure*}
The simulator captures system patterns from real data and can flexibly reflect the state changes of the system under decision-making. 
However, simple observation is not enough to support the need to reduce decision risks in a traceable way in the supply chain, so we propose a decision-maker model in this section, which elevates the observation of the data set to the level of intervention. 
By modifying the strategy in a targeted manner and observing the counterfactual scenarios generated by the simulator, we can finally obtain a highly capable decision-maker model.
Specifically, as shown in the Figure \ref{fig:dm}, our decision-makers combine the experience of historical decisions and the estimation of future benefits to give the final optimal decision. With the linkage to the simulator, the decision-makers have more experience and a clearer estimation of the benefits of the decision, and gradually iterate to learn better decision strategies.

\noindent \textbf{Decision Network.}
We construct a learnable decision-making network $\mathcal{M}$ to generate the probability of each potential shipping mode for each order $n$:
\begin{equation}
\mathbf{\hat{R}}_n = \mathcal{M}(F^I_n; \Phi) = (R^1_n, \dots, R^{|F^D|}_n), 
\end{equation}
where $R^{d}_n$ denotes the probability to choose the $d^{th}$ shipping mode.

\noindent\textbf{Historical Experience.}
History contains rich experience, and we give decision networks the ability to learn from history.
We encourage the selection of the shipping mode with the highest expected utility based on historical data.  
For each order event $n$, the decision-maker observes the system state $F^I_n$ and outputs a probability distribution $\mathbf{\hat{R}}_n$.  
A decision $\hat{f}^D_n$ is then sampled from this distribution using the Gumbel-Softmax reparameterization method~\cite{gumbel-softmax} to ensure the entire process is differentiable.  

To evaluate each possible decision, we compute the expected values of $T^{Time}$ and $T^{Profit}$ for that decision based on historical data. The reward $E^{d}_n(f_d^D)$ for taking decision $f_d^D$ is defined as:
\begin{equation}
E^{d}(f_d^D) = \mathbb{E}[T^{Time} | f_d^D] + \mathbb{E}[T^{Profit} | f_d^D].
\end{equation}
where $\mathbb{E}[T^{Time} | f_d^D]$ and $\mathbb{E}[T^{Profit} | f_d^D]$ are the expectation of $T^{Time}$ and $T^{Profit}$ when choose shipping mode $f_d^D$ calculated from historical data.
The decision network $\mathcal{M}$ is trained using the loss function, $\mathcal{L}_{\text{h}}$, which encourages selecting decisions with the highest expected reward:
\begin{equation}
\mathcal{L}_{\text{h}} = -\sum_{n=1}^{N} \sum_{d=1}^{|F^D|} \mathbf{\hat{R}}^d_{n} \cdot E^{d}(f_d^D),
\end{equation}
where $\hat{R}^d_{n}$ is the probability of selecting decision $f^D_d$ for order $n$, obtained from the probability distribution $\hat{R}^d_n= \mathcal{M}(F^I_n;\Phi)$. 
This loss formulation directly optimizes the parameters $\Phi$ of $\mathcal{M}$, training the network to prioritize the shipping mode that maximizes the expected combined $E^{d}(f_d^D)$ from historical experience. 

\noindent
\textbf{Future Estimation.}
In addition to learning from experience, having a foresight into the future is also necessary for making good decisions. We model the decision-making task as a contextual bandit problem~\cite{contextualBandits} to estimate the future chosen shipping mode of each order. 
This approach is compatible with scenarios where the computation of key measures is non-differentiable, such as retrieving profit or other system-level metrics from historical data. 

The decision network $\mathcal{M}$ can be regarded as a value network that estimates the overall batch reward resulting from a particular shipping mode applied to an order. For each order $n$, the chosen shipping mode also sampled from the distribution $\mathbf{\hat{R}}_{n}$ which can be understood as a normalized reward score:
\begin{equation}
\hat{f}^D_{d, n} = \text{GumbelSoftmax}(\sigma(\mathcal{M}(F^I_n;\Phi))),
\end{equation}
Herein, $\hat{f}^D_{d, n}$ denotes the shipping mode chosen for order $n^{th}$ is $d^{th}$ shipping mode. 
For each order $n$, the simulator predicts statuses $\hat{f}^E_{e,n}$ based on the inherent attributes $F^I_n$ and the selected decision $\hat{f}^D_{d, n}$: 
\begin{equation} 
\hat{F}^E_n = \mathcal{S}(F^I_n, \hat{f}^D_{d,n}). 
\end{equation} 
The simulator is pre-trained, with its parameters frozen during decision-making. This design ensures that the simulator provides a stable and reliable risk-free environment, allowing us to freely explore and evaluate a wide range of decisions without concern for real-world consequences. 
The orders are grouped according to the selected shipping mode $\hat{f}^D_{d,n}$. We calculate the timely delivery rate and profit for each shipping mode using the necessary information, including order information, selected shipping mode, and the simulated order status.
We use $T_{d}^{Time}$ and $T_{d}^{Profit}$ to denote the timely delivery rate and the profit for those orders that choose the shipping mode $f_{d}^D$.
The total reward for shipping mode $f_d^D$ in a batch is computed by:
\begin{equation}
R^d_{\text{batch}} = T_{d}^{Time} + T_{d}^{Profit}.
\end{equation}

The value network is trained to minimize the prediction error of batch rewards by optimizing the following loss:
\begin{equation}
\mathcal{L}_{\text{f}} = \frac{1}{N} \sum_{n=1}^{N}\sum_{d=1}^{|F^D|} \Big(\hat{R}^d_{n} - R^d_{\text{batch}}\Big)^2.
\end{equation}

Through iterative feedback, the decision-maker network refines its understanding of the relationship between individual decisions and their impact on batch-level performance. 

\noindent
\textbf{Combined Loss Function and Decision-Making.}
To ensure that the model can benefit from both historical experience and future estimates, the overall loss function combines the objectives of the two levels:
\begin{equation}
\mathcal{L}_\mathcal{M} =\mathcal{L}_{\text{f}} + \lambda \cdot \mathcal{L}_{\text{h}},
\end{equation}
where $\lambda$ is the hyperparameter that balances the importance of historical experience and future predictions.

Through the integration of two optimization objectives, we develop a multi-perspective decision network $\mathcal{M}$. This network effectively combines lessons learned from history with visionary predictions for the future.
Given the inherent attributes $F^I_n$ for each order $n$, $\mathcal{M}$ outputs a value for each potential decision $f_d^D$, denoted as $\hat{R}_{n}=\mathcal{M}(F^I_n;\Phi)$. This value can be interpreted as either the probability or the reward associated with each shipping mode. 
The final decision $\hat{f}_d^D$ is determined by selecting the decision corresponding to the maximum value output by the network:
\begin{equation}
\hat{f}_d^D = \arg\max_{f_d^D \in F^D} \mathcal{M}(F^I_n;\Phi).
\end{equation}

\section{Experiments}

\subsection{Experimental Setting}
\noindent \textbf{Datasets Description.}  
We conducted experiments on three real-world supply chain datasets: DataCo~\cite{DataCo}, Global-Store~\cite{GS}, and OAS~\cite{OAS}. These datasets cover different logistics and transportation settings, including various shipping modes. Each dataset includes detailed order records, such as order information, selected shipping modes, and delivery statuses, providing a realistic testbed for evaluating \shortname.  
To ensure a fair and consistent evaluation, each dataset is randomly divided into training, validation, and test sets in an 8:1:1 ratio. This split helps mitigate potential temporal biases and improve generalization by avoiding overfitting to chronological patterns. To further reduce overfitting risk, we apply standard regularization techniques, including early stopping based on validation loss and $\ell_2$ weight penalties during training.
Both the simulator and the decision network are trained only on the training set, and the simulation performance and post-decision metrics on the test set are reported. Each experiment is repeated five times with different random seeds, and we report average results. The dataset statistics are summarized in the first row of Table~\ref{tab:sim_overall} in the form (\#Features, \#Instances).


\begin{table*}
\centering
\caption{\small{Comparison of simulation results. Overall represents the average accuracy of all evolutionary features.}}\label{tab:sim_overall}
\renewcommand{\arraystretch}{1.1} 
\resizebox{\textwidth}{!}{
\begin{tabular}{c|cccc|cccc|cccc} 
\toprule
Dataset & \multicolumn{4}{c|}{Dataco (43, 165445)} & \multicolumn{4}{c|}{GlobalStore (27, 51290)} & \multicolumn{4}{c}{OAS (22, 28136)} \\ 
\hline
Method & Markov & Prediction & Generation & \cellcolor{gray!20}Sim-to-Dec 
       & Markov & Prediction & Generation & \cellcolor{gray!20}Sim-to-Dec 
       & Markov & Prediction & Generation & \cellcolor{gray!20}Sim-to-Dec \\ 
\hline\hline
$f^E_{\text{risk}}$    
& 0.4978 & 0.7019 & \uline{0.7024} & \cellcolor{gray!20}\textbf{0.9508} 
& 0.4961 & 0.8440 & \uline{0.9366} & \cellcolor{gray!20}\textbf{0.9743} 
& 0.5100 & \uline{0.7157} & 0.7149 & \cellcolor{gray!20}\textbf{0.7215} \\
$f^E_{\text{time}}$    
& 0.1487 & 0.3395 & \uline{0.3485} & \cellcolor{gray!20}\textbf{0.8851} 
& 0.1355 & 0.6767 & \uline{0.8066} & \cellcolor{gray!20}\textbf{0.9255} 
& 0.0011 & 0.3706 & \uline{0.3916} & \cellcolor{gray!20}\textbf{0.3985} \\
$f^E_{\text{status}}$  
& 0.5040 & \uline{0.8161} & 0.8156 & \cellcolor{gray!20}\textbf{0.9695} 
& 0.4934 & 0.8430 & \uline{0.9355} & \cellcolor{gray!20}\textbf{0.9756} 
& 0.5068 & \uline{0.7510} & 0.7503 & \cellcolor{gray!20}\textbf{0.7574} \\
Overall $\uparrow$    
& 0.3835 & 0.6191 & \uline{0.6221} & \cellcolor{gray!20}\textbf{0.9351} 
& 0.3750 & 0.7879 & \uline{0.8929} & \cellcolor{gray!20}\textbf{0.9585} 
& 0.3393 & 0.6124 & \uline{0.6189} & \cellcolor{gray!20}\textbf{0.6258} \\
\bottomrule
\end{tabular}}
\end{table*}

\noindent \textbf{Evaluation Metrics.}  
For the simulator, we evaluate its accuracy by comparing the predicted order status attributes (delay risk, delivery time, and on-time status) against the ground-truth values in the test set. Accuracy is computed for each attribute individually, and an overall accuracy is reported as the unweighted average across the three prediction tasks. This provides a holistic measure of the simulator’s fidelity.
For the decision-maker, we compare the average profit and on-time rate of test set orders. We normalize both indicators to the [0, 1] range, and report two aggregate metrics: the absolute difference between the two objectives (Diff), and their sum (Overall), to comprehensively assess decision quality.


\noindent \textbf{Baseline Algorithms.}
To evaluate simulation methods, we selected three paradigms to simulate the supply chain:  
(1) \textit{Markov-based simulation}~\cite{markov}, representing traditional approaches that use state-transition probabilities to model system dynamics;  
(2) \textit{Prediction-based simulation}~\cite{mlp-based}, which adopts a multi-task framework to predict evolutionary features individually based on input conditions; and  
(3) \textit{Non-autoregressive generation-based simulation}~\cite{gen-based}, which generates multiple evolutionary features simultaneously in one step. For simplicity, we refer to this paradigm as \textit{Generation} in the following sections.
For decision-making based on simulation, we selected three paradigms:  
(1) \textit{Linear Programming (LP)}~\cite{LP}, a traditional optimization method for solving predefined decision problems;  
(2) \textit{Reinforcement Learning}, where a DQN-based RL agent iteratively optimizes strategies through interaction with the environment; and  
(3) \textit{LLM-based decision-making}~\cite{LLM-based}, which leverages the expert knowledge embedded in ChatGPT-3.5, a large language model (LLM), to make decisions under a zero-shot setting without task-specific fine-tuning.





\subsection{Experimental Results}

\noindent\textbf{A Study of Generative Simulator Accuracy.}
As shown in Table \ref{tab:sim_overall}, we compare our model with other baselines on three real-world supply chain datasets. We have the following observations:
(1) \shortname~ outperforms all baseline methods on all datasets. Specifically, in terms of overall accuracy, our method improves the strongest baseline by 50.3\%, 7.3\%, and 1.1\% in DataCo, ClobalStore and OAS respectively. We attribute this to our unique insight that enables the simulator to sequentially reproduce realistic changes to the system at a fine-grained level.
(2) Compared with simulation methods based on Markov chains, data-driven simulation has improved accuracy. For example, on the DataCo dataset, prediction and generation and \shortname~ have improved 61.4\%, 62.2\% and 143.8\% respectively. This shows that data-driven methods can mine the potential laws in the data to better model the system operation.
\begin{table}[ht]
\centering
\caption{\small{Comparison of decision-making results (transposed). Each row corresponds to a method; each dataset shows four metrics.}}\label{tab:dm_overall}
\renewcommand{\arraystretch}{1}
\small
\resizebox{\linewidth}{!}{
\begin{tabular}{c|cccc|cccc|cccc}
\toprule
\multirow{2}{*}{Method} 
  & \multicolumn{4}{c|}{\textbf{DataCo}} 
  & \multicolumn{4}{c|}{\textbf{Global-Store}} 
  & \multicolumn{4}{c}{\textbf{OAS}} \\
\cline{2-13}
  & $T^{\text{Time}} \uparrow$ & $T^{\text{Profit}} \uparrow$ & $\text{Diff} \downarrow$ & $\text{Overall} \uparrow$
  & $T^{\text{Time}} \uparrow$ & $T^{\text{Profit}} \uparrow$ & $\text{Diff} \downarrow$ & $\text{Overall} \uparrow$
  & $T^{\text{Time}} \uparrow$ & $T^{\text{Profit}} \uparrow$ & $\text{Diff} \downarrow$ & $\text{Overall} \uparrow$
  \\
\hline\hline
Real 
  & 0.5244 & 0.0364 & 0.4880 & 0.5608
  & 0.3320 & 0.0848 & \textbf{0.2472} & 0.4168
  & 0.4800 & 0.0000 & 0.4800 & 0.4800 \\
LP 
  & 0.5162 & 0.5434 & \uline{0.0272} & \uline{1.0596}
  & 0.3552 & 0.6001 & \uline{0.2449} & 0.9554
  & 0.5037 & 0.1043 & \uline{0.3994} & \uline{0.6080} \\
RL 
  & 0.5276 & 0.2071 & 0.3205 & 0.7347
  & 0.2827 & 0.9326 & 0.6499 & \uline{1.2153}
  & 0.4817 & 0.0000 & 0.4817 & 0.4817 \\
LLM 
  & 0.5258 & 0.2459 & 0.2800 & 0.7717
  & 0.3298 & 0.0439 & 0.2859 & 0.3736
  & 0.4844 & 0.0000 & 0.4844 & 0.4844 \\
\rowcolor{gray!15}
Sim-to-Dec 
  & 0.5397 & 0.5637 & \textbf{0.0240} & \textbf{1.1034}
  & 0.3446 & 0.9278 & 0.5828 & \textbf{1.2724}
  & 0.4882 & 0.1611 & \textbf{0.3271} & \textbf{0.6493} \\
\bottomrule
\end{tabular}
}
\end{table}



\begin{figure*}
  \centering
  \subfloat[\centering \small{Training\&Test}]{
    \includegraphics[width=0.19\textwidth]{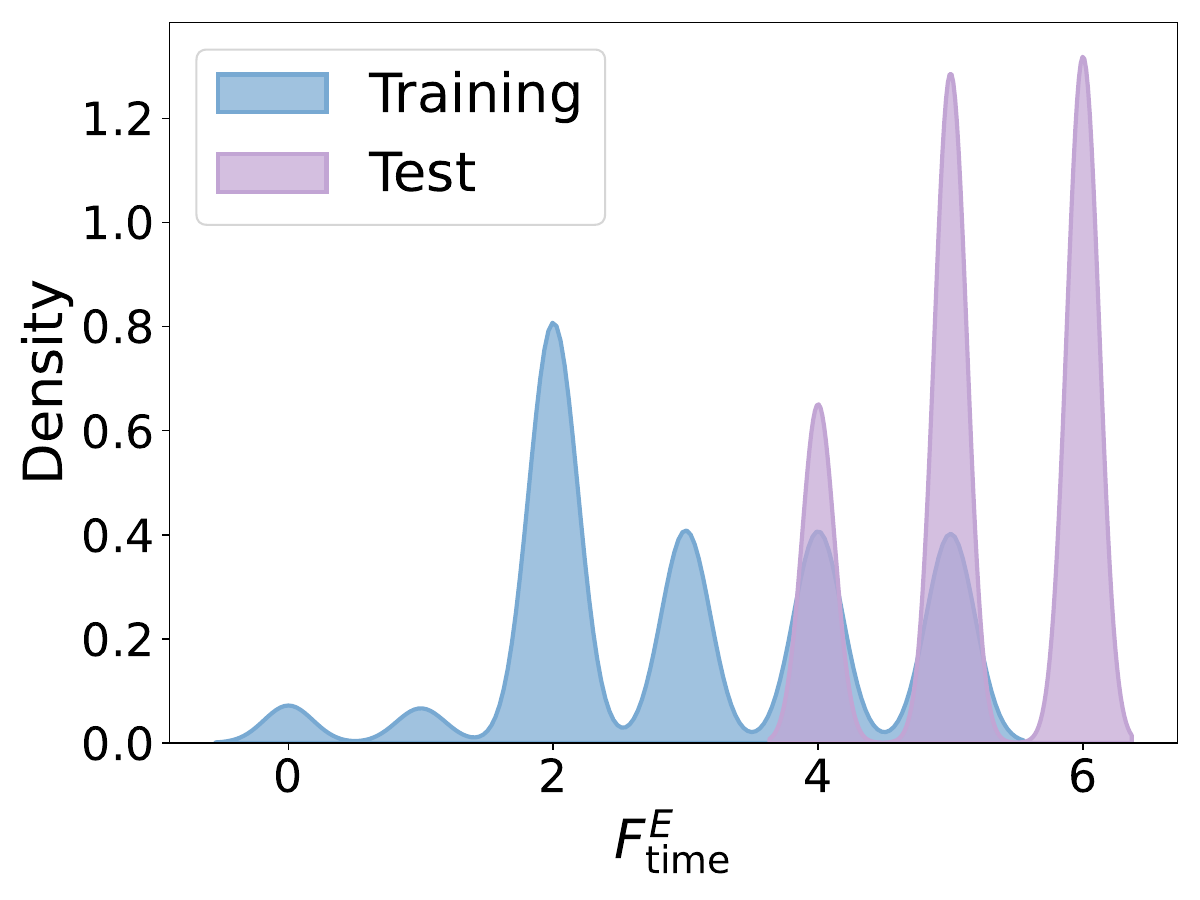}
  }
    \subfloat[\centering \small{Sim-to-Dec}]{
    \includegraphics[width=0.19\textwidth]{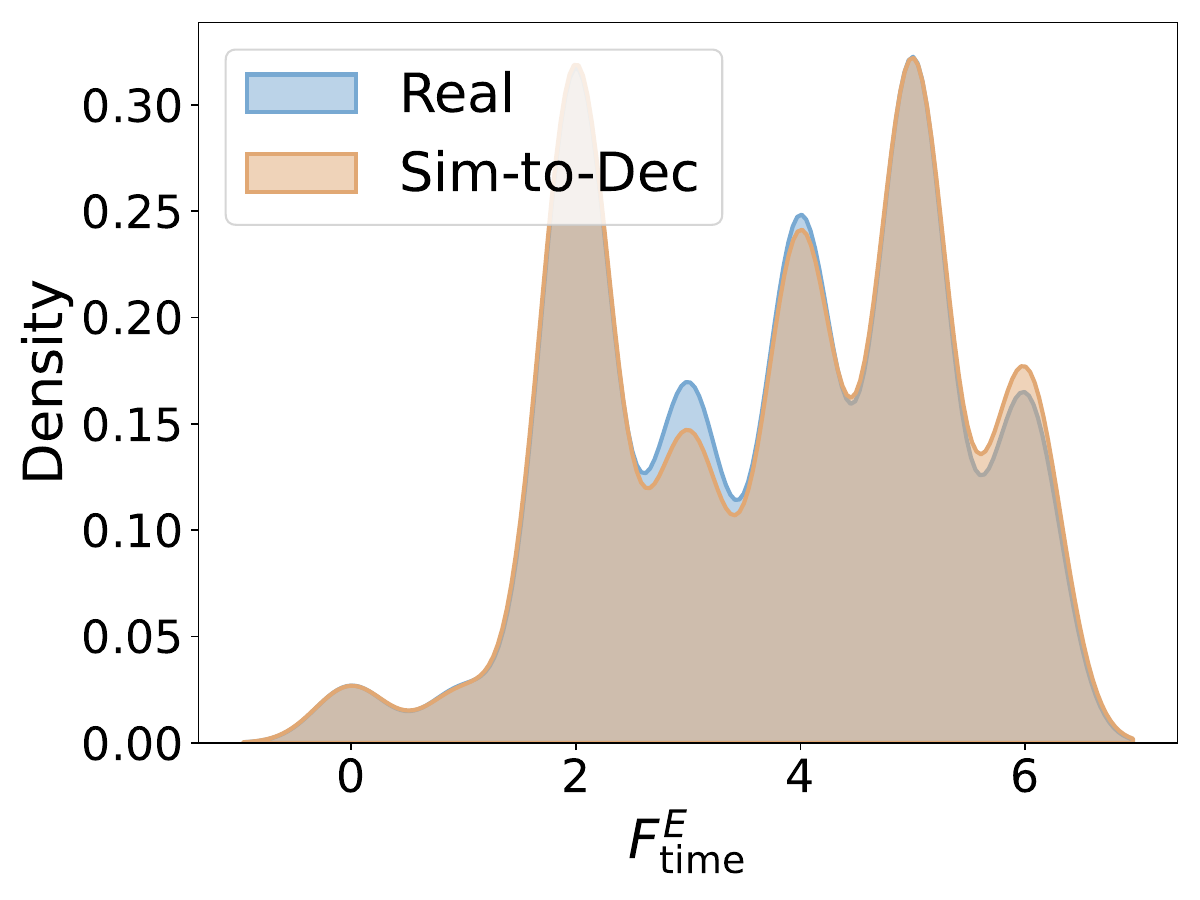}
  }
    \subfloat[\centering \small{Prediction}]{
    \includegraphics[width=0.19\textwidth]{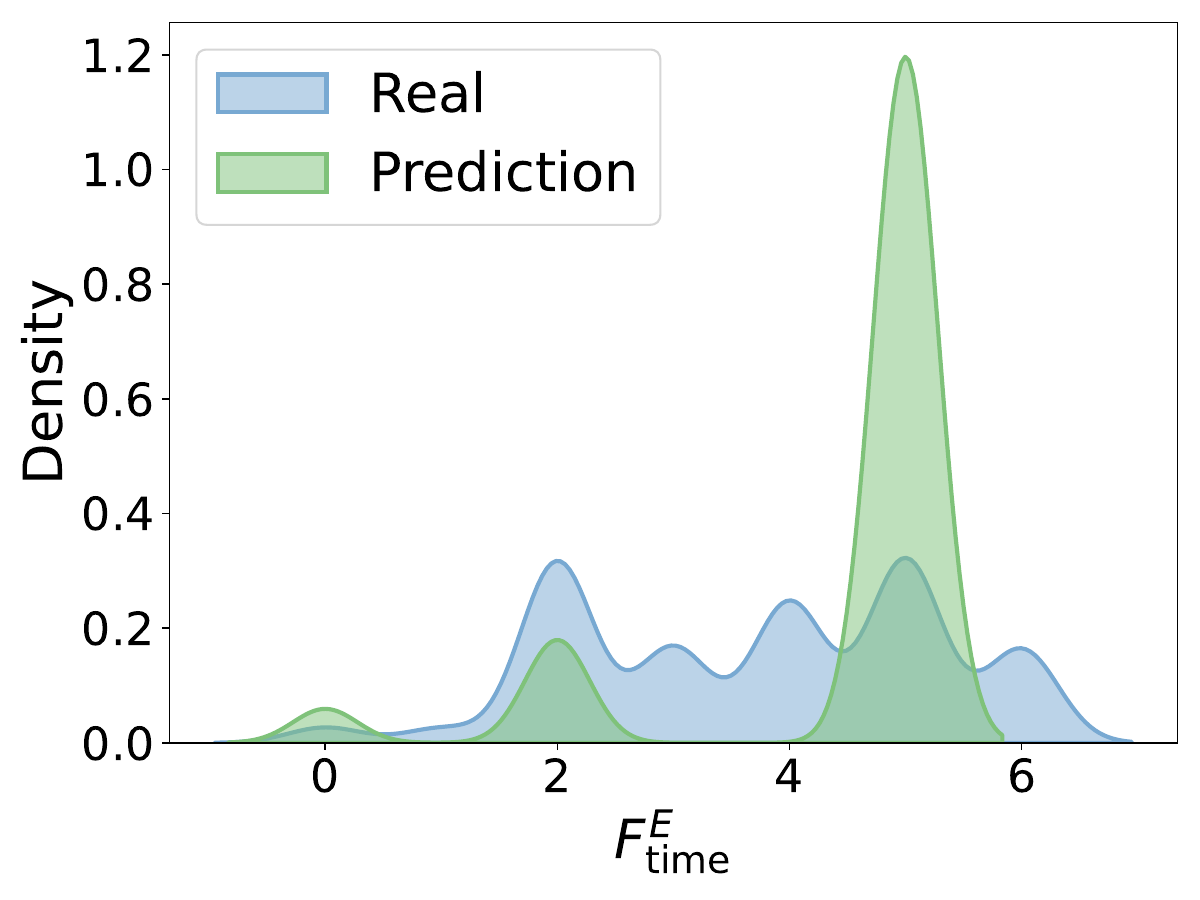}
  }
  \subfloat[\centering \small{Generation}]{
    \includegraphics[width=0.19\textwidth]{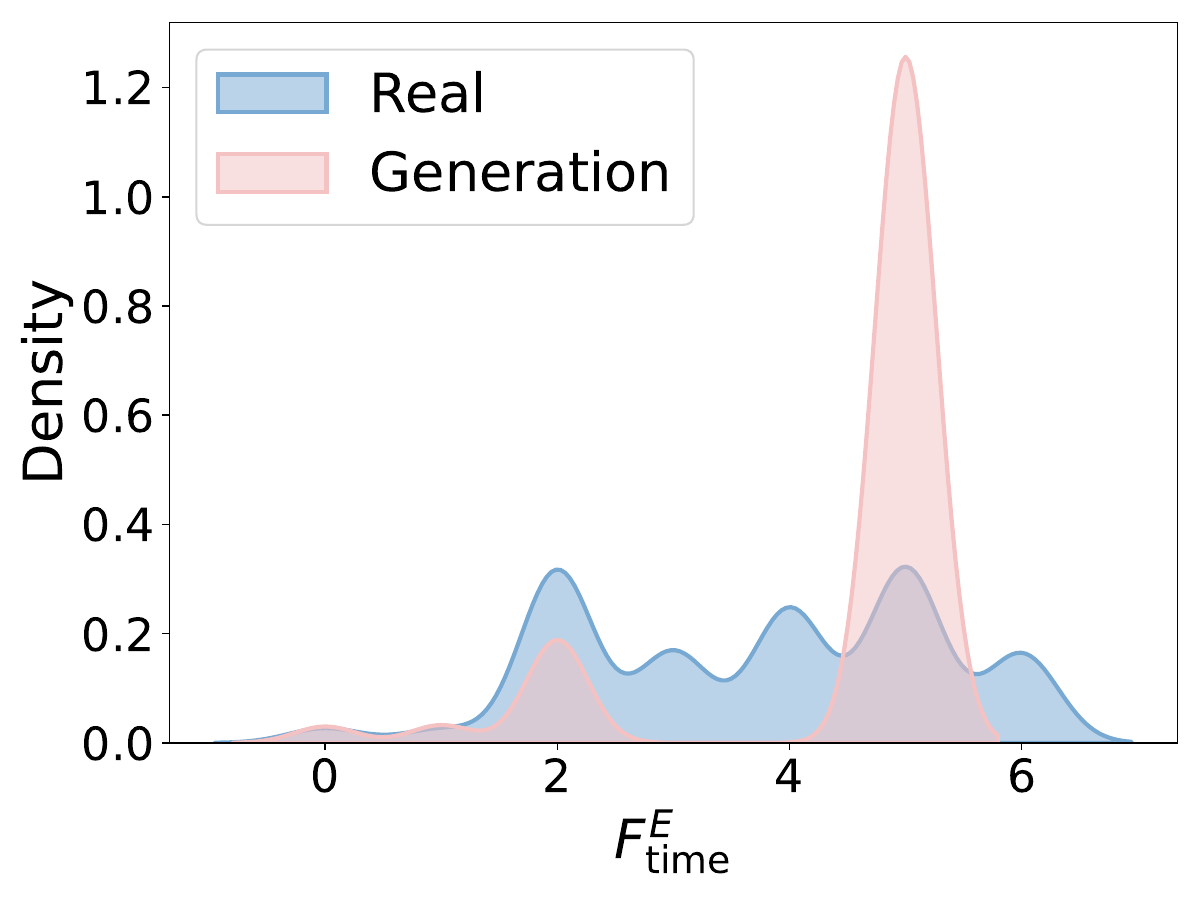}
  }
\subfloat[\centering \small{Markov}]{
    \includegraphics[width=0.19\textwidth]{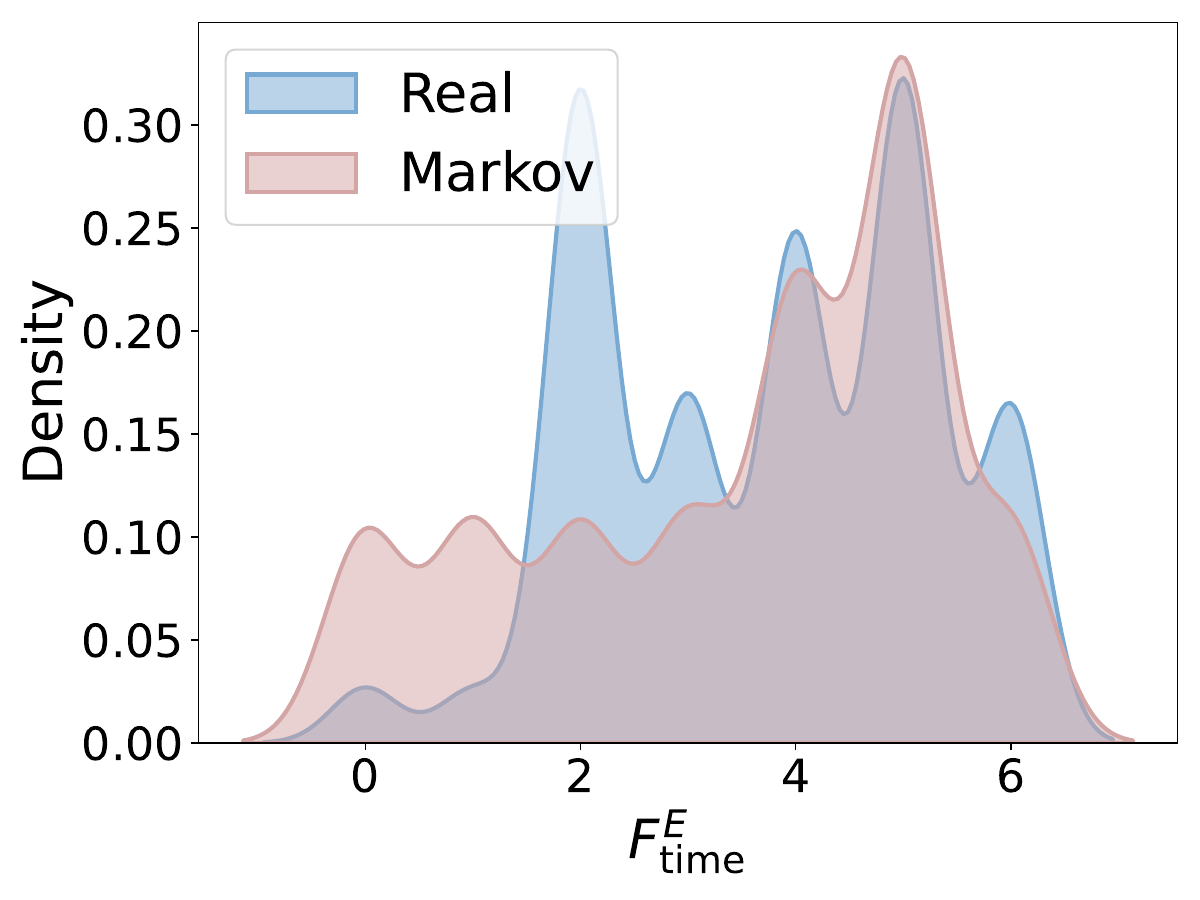}
  }
\caption{\small{Simulation under distribution shift on DataCo dataset}}\label{fig:ood_dis}
\end{figure*}

\noindent\textbf{Robustness Check of Generative Simulator against Distribution Shift.}
The dynamic complexity of the real world means that the environment in which the system operates is constantly changing. Mining potential patterns from data and being able to accurately simulate when the operating environment changes or is disturbed is an ideal property of a robust simulator.
We designed an experiment to verify the performance of our method in this scenario. 
We repartitioned the DataCo dataset according to $f^E_{\text{time}}$, so that the data distribution of the training set and the test set shifted, as shown in Figure \ref{fig:ood_dis}(a). 
We performed simulations in this changing environment. As shown in Figure \ref{fig:ood_dis}(b)-(e), we compared the $f^E_{\text{time}}$ distributions generated by different simulation methods with the distribution in the real test set. The experimental results show that our method can still capture the underlying patterns of the data when the distribution shifts, and deduce evolutionary variables that conform to the actual situation, verifying the practicality of our method.

\noindent\textbf{A Study of Decision-Maker.}
We first compare the performance of different decision-makers when improving key metrics at the same time. As shown in Table \ref{tab:dm_overall}, we have the following observations:
(1) Our method shows the best performance when considering both the overall measure (Overall) and the gap between the two measures (Diff). For example, compared with the strongest baseline, our method improves on oveall by  4.1\% and 6.8\%, and the Diff is reduced by 11.8\% and 18.1\% on the DataCo and OAS datasets. This shows that the decision-maker we proposed has strong decision-making ability and can balance conflicting optimization goals to achieve common improvement.
(2) LP performs well on some datasets. However, its performance is achieved by enumerating all possible strategies and their associated rewards on the test set, rather than learning a generalizable policy during training. While this approach leverages more task-specific information, it significantly limits the method's generalization capability and results in substantial computational overhead, making it impractical for large-scale scenario.

\subsection{In-depth Analysis: Ablation Studies, Parameter Sensitivity, Computational Efficiency}

\begin{figure}
  \centering
  \subfloat[\centering \small{Decision-Maker Ablation on DataCo}]{
    \includegraphics[width=0.23\textwidth]{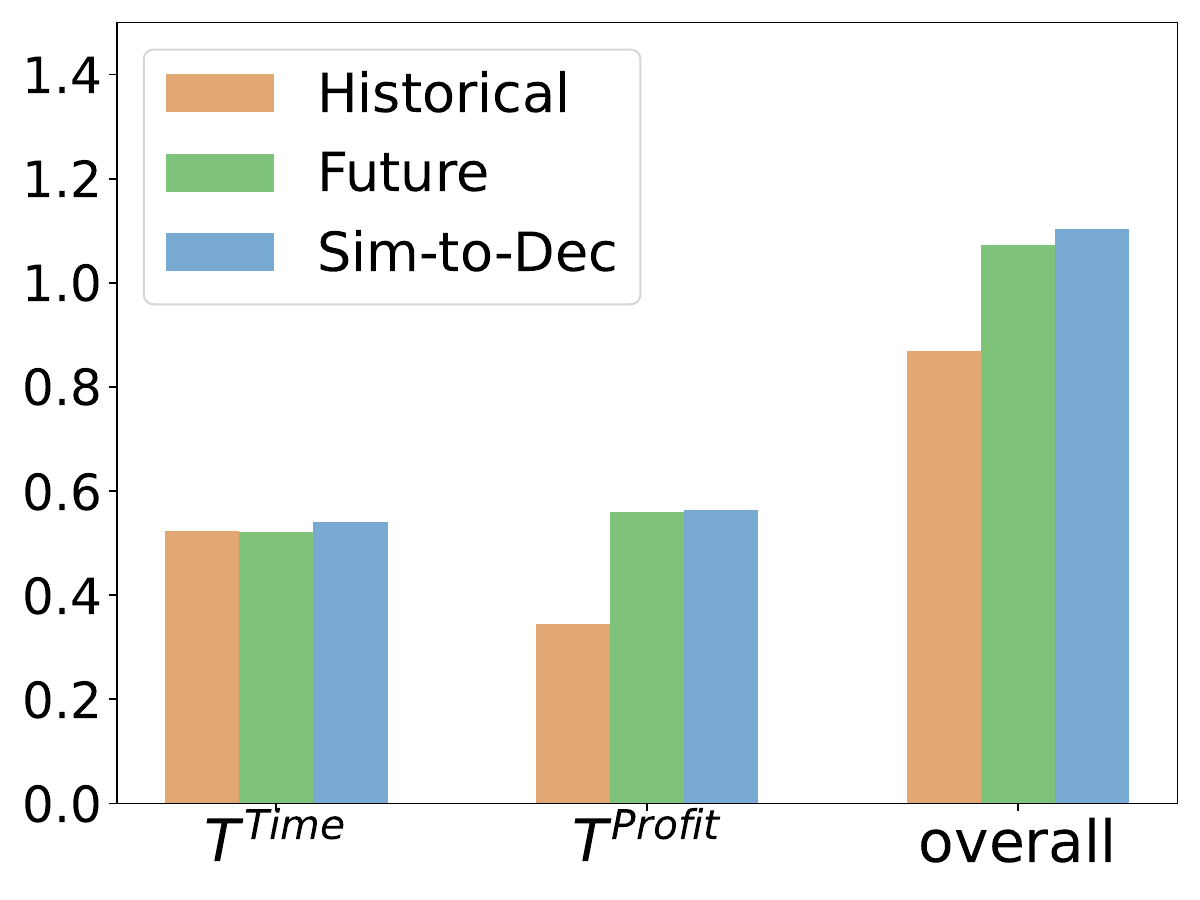}
  }
    \subfloat[\centering \small{Decision-Maker Ablation on GlobalStore}]{
    \includegraphics[width=0.23\textwidth]{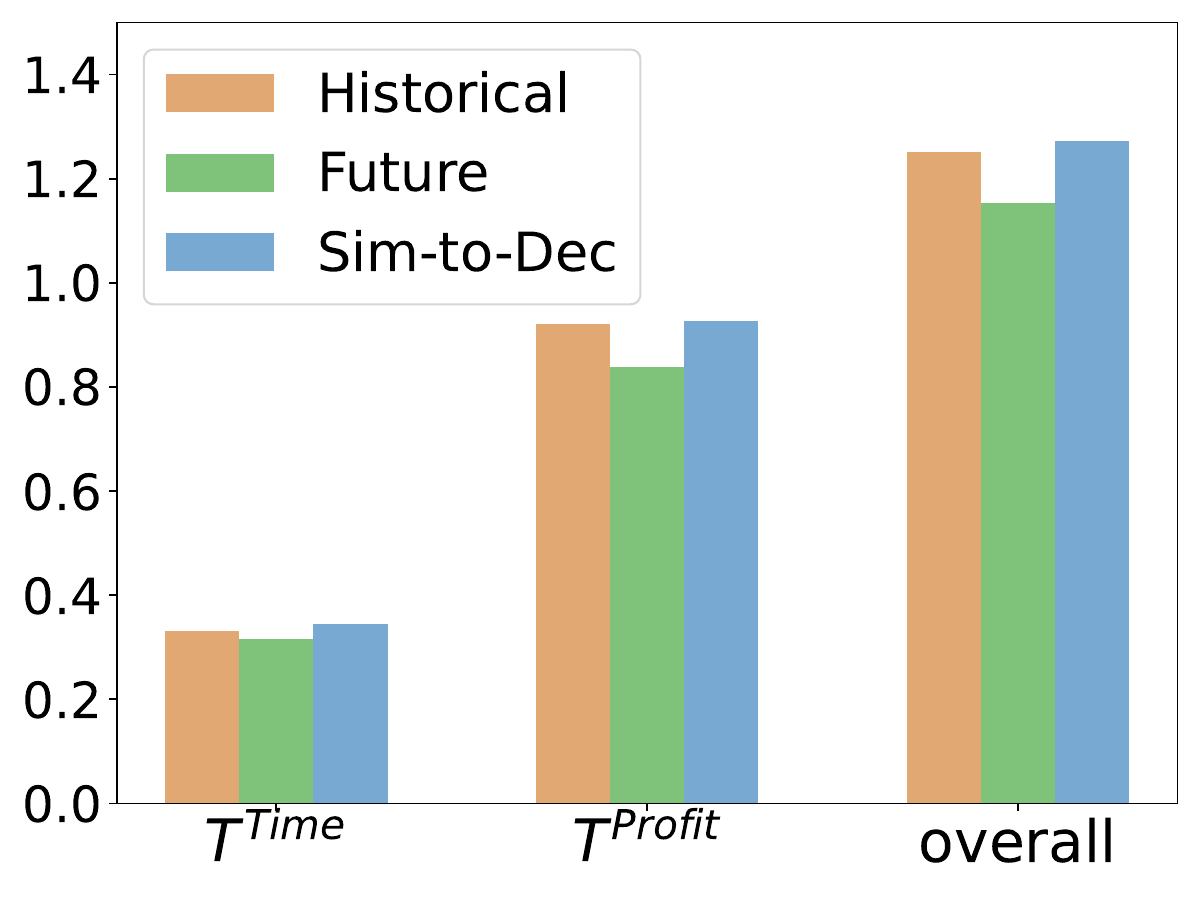}
  }
    \subfloat[\centering \small{Sensitivity Analysis on DataCo}]{
    \includegraphics[width=0.23\textwidth]{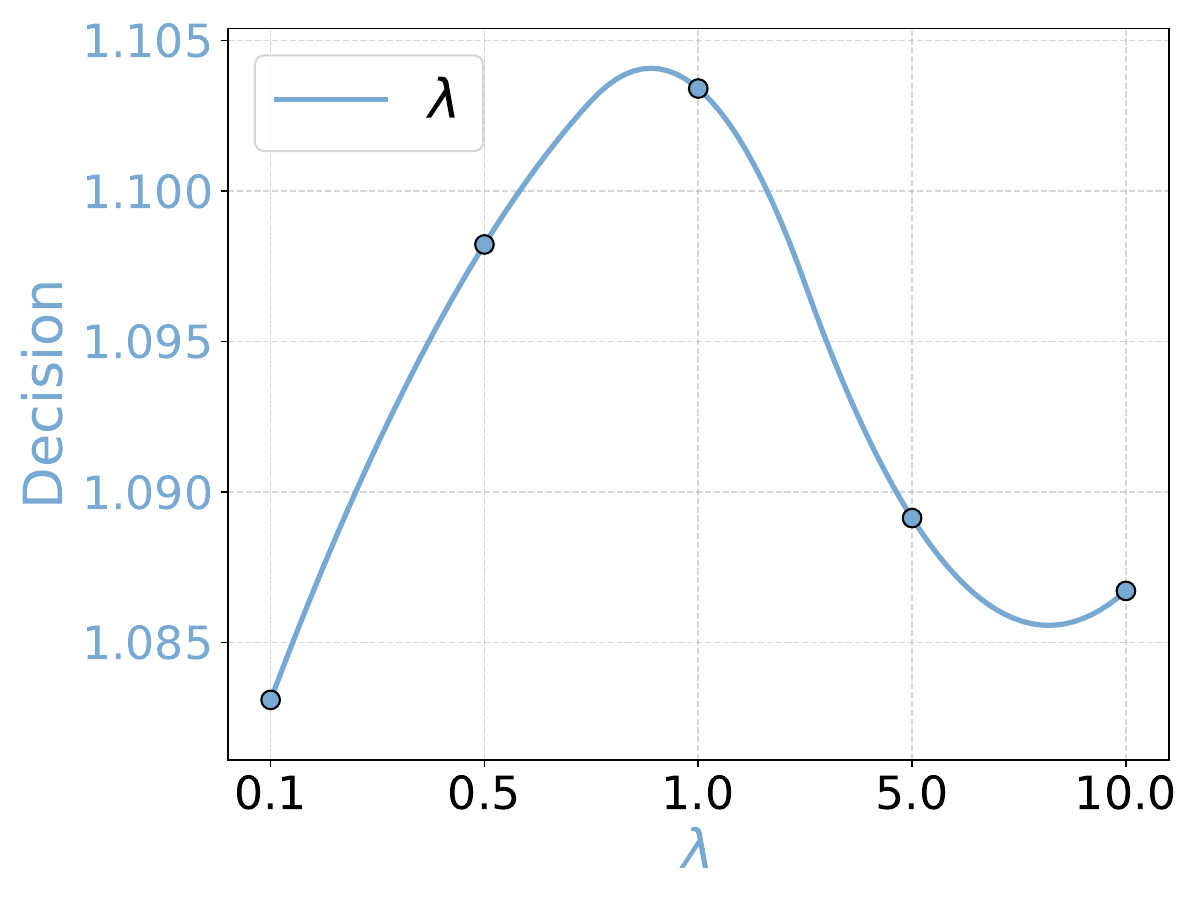}
  }
  \subfloat[\centering \small{Sensitivity Analysis on GlobalStore}]{
    \includegraphics[width=0.23\textwidth]{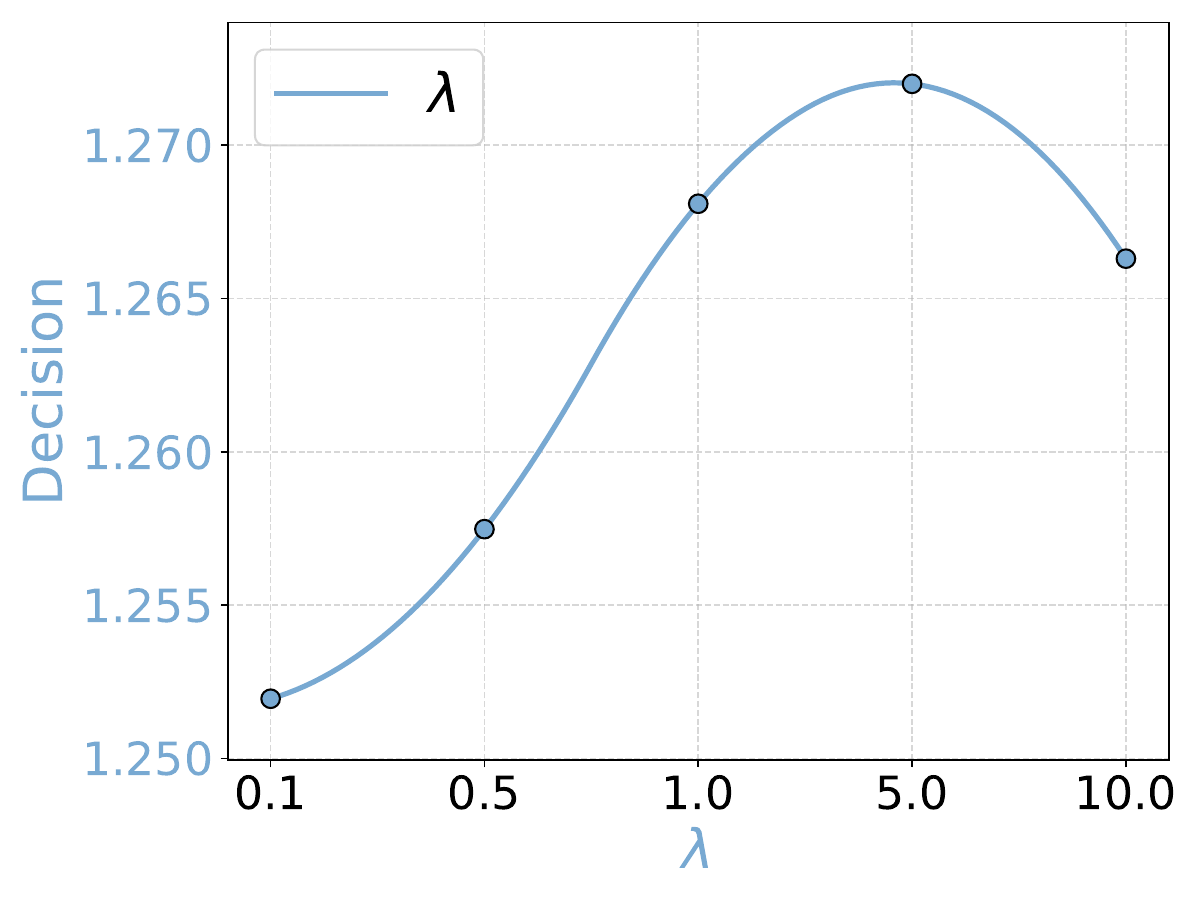}
  }
\caption{\small{Ablation Study and Sensitivity Analysis}}\label{fig:absa}
\end{figure}
\begin{wraptable}{r}{0.4\textwidth}
\scriptsize
\centering
\caption{\small{Time Cost of Simulator Training}}
\label{tab:time_sim}
\resizebox{0.4\textwidth}{!}{
\begin{tabular}{c|c|ccc} 
\toprule
Dataset & Method & Time/Epoch (s) & \# Epoch & Total Time (s) \\ 
\midrule
\multirow{4}{*}{Dataco} 
    & Markov     & -    & -   & 53  \\
    & Prediction & 0.86 & 60  & 70  \\
    & Generation & 0.74 & 110 & 81  \\
    & Sim-to-Dec       & 0.90 & 350 & 315 \\ 
\midrule
\multirow{4}{*}{GlobalStore} 
    & Markov     & -    & -   & 14  \\
    & Prediction & 0.12 & 210 & 25  \\
    & Generation & 0.19 & 200 & 38  \\
    & Sim-to-Dec       & 0.24 & 200 & 48  \\
\bottomrule
\end{tabular}}
\end{wraptable}
We evaluate the time complexity of our framework and baseline methods. Tables~\ref{tab:time_sim} shows that our simulator demonstrates competitive training efficiency. Compared with lightweight but less expressive baselines like Markov models, our method achieves fine-grained learning in a reasonable time, balancing accuracy and runtime cost.
Figure \ref{fig:absa}(a) and \ref{fig:absa}(b) show that decisions based on historical experience and future estimates work differently on different data sets, but a combination of the two usually leads to better decisions.
As shown in Figure \ref{fig:absa}(c) and \ref{fig:absa}(d), we use $\lambda$ to balance the proportion of historical experience and future estimates in decisions. As $\lambda$ increases, the performance first increases and then decreases, indicating that a suitable $\lambda$ should be selected so that the decision can benefit from both perspectives.
For decision-making, our method requires only 1.2 seconds on the DataCo dataset and 0.2 seconds on Global-Store, significantly faster than the LLM-based approach (1856s and 655s). LP methods take 8s and 2s respectively, while RL is fastest (0.5s and 0.1s) but less effective in decision quality. Overall, our approach offers a strong trade-off between efficiency and performance.

\section{Conclusion}
In this work, we propose a unified framework that tightly integrates a \textit{generative simulator} with a \textit{feedback-driven decision-maker} to improve responsiveness in supply chain transportation. The simulator models order dynamics through autoregressive learning, enabling fine-grained prediction of shipment evolution under different transportation strategies. The decision-maker iteratively refines shipping mode selection by combining historical patterns with forward-looking reward estimation, guided by simulated feedback. This tight simulation–decision coupling overcomes the limitations of static models and manual heuristics, providing enhanced flexibility and adaptability in dynamic logistics environments.
Extensive experiments on real-world supply chain datasets demonstrate the superiority of our approach in balancing timely delivery and cost efficiency, even under distribution shifts. These results underscore the potential of our framework for broader applications in logistics optimization and adaptive decision-making systems.

\section*{ACKNOWLEDGMENTS}
This work was supported by the U.S. National Science Foundation under Grant Nos. 2426340, 2416727, 2421864, 2421865, and 2421803. The authors thank all members of the research group for their valuable feedback and insightful discussions during the development of this work.

\footnotesize

\bibliographystyle{wsc}

\bibliography{references}

\section*{AUTHOR BIOGRAPHIES}
\noindent {\bf \MakeUppercase{Haoyue Bai}} is a PhD student at Arizona State University. His email address is \email{baihaoyue621@gmail.com}.\\
\noindent {\bf \MakeUppercase{Haoyu Wang}}  is a researcher in the DSSS department at NEC Labs America. His email address is \email{haoyu@nec-labs.com}.\\
\noindent {\bf \MakeUppercase{Nanxu Gong}} is a PhD student at Arizona State University. His email address is \email{nanxugong@outlook.com}.\\
\noindent {\bf \MakeUppercase{Xinyuan Wang}} is a PhD student at Arizona State University.  His email address is \email{wangxinyuan0921@gmail.com}.\\
\noindent {\bf \MakeUppercase{Wangyang Ying}} is a PhD student at Arizona State University. His email address is \email{yingwangyang@gmail.com}.\\
\noindent {\bf \MakeUppercase{Haifeng Chen}} is the head of the DSSS at NEC Laboratories America. His email address is \email{haifeng@nec-labs.com}.\\
\noindent {\bf \MakeUppercase{Yanjie Fu}} is an associate professor at Arizona State University. His e-mail address is \email{yanjie.fu@asu.edu}.\\

\end{document}